\documentclass[conference]{IEEEtran}
\IEEEoverridecommandlockouts
\usepackage{cite}
\usepackage{amsmath,amssymb,amsfonts}
\usepackage{algorithmic}
\usepackage{graphicx}
\usepackage{textcomp}
\usepackage{xcolor}
\usepackage{booktabs}
\usepackage{tabularx}
\usepackage{multirow}
\usepackage{microtype}
\usepackage{fontspec}
\newfontfamily\bnfont{NotoSansBengali}[
  Path = fonts/,
  Extension = .ttf,
  UprightFont = *-Regular,
  BoldFont = *-Regular,
  ItalicFont = *-Regular,
  BoldItalicFont = *-Regular,
  Script = Bengali
]
\newcommand{\bn}[1]{{\bnfont #1}}
\usepackage{hyperref}
\hypersetup{colorlinks=true,linkcolor=black,citecolor=black,urlcolor=blue}
\usepackage{fancyhdr}

\fancypagestyle{firstpage}{%
  \fancyhf{}
  \fancyhead[R]{\scriptsize\itshape 11th IEEE Asia-Pacific Conference on Computer Science and Data Engineering (IEEE CSDE 2026)\\1-3 November 2026, Cox's Bazar, Bangladesh}

}

\begin{document}

\title{PhysicsMate: A Curriculum-Grounded Bengali Benchmark for Secondary Physics QA with Small-Model Adaptation}

\author{
\IEEEauthorblockN{1\textsuperscript{st} Rashid Azraf Jahin}
\IEEEauthorblockA{\textit{Dept. of Electrical and Computer Engineering} \\
\textit{North South University}\\
Dhaka, Bangladesh \\
rashid.jahin@northsouth.edu}
\and
\IEEEauthorblockN{2\textsuperscript{nd} Saadman Sajid}
\IEEEauthorblockA{\textit{Dept. of Electrical and Computer Engineering} \\
\textit{North South University}\\
Dhaka, Bangladesh \\
saadman.sajid@northsouth.edu}
\and
\IEEEauthorblockN{3\textsuperscript{rd} Khan Raiyan Ibne Reza}
\IEEEauthorblockA{\textit{Dept. of Electrical and Computer Engineering} \\
\textit{North South University}\\
Dhaka, Bangladesh \\
raiyan.reza@northsouth.edu}
\and
\IEEEauthorblockN{4\textsuperscript{th} Sumaiya Tabassum Nimi}
\IEEEauthorblockA{\textit{Dept. of Electrical and Computer Engineering} \\
\textit{North South University}\\
Dhaka, Bangladesh \\
sumaiya.nimi@northsouth.edu}
}

\IEEEpubid{\makebox[\columnwidth]{979-8-3195-4477-3/26/\$31.00~\copyright~2026~IEEE\hfill}\hspace{\columnsep}\makebox[\columnwidth]{}}

\maketitle
\thispagestyle{firstpage}

\begin{abstract}
Bengali secondary education lacks curriculum-grounded benchmarks for STEM question-solving, and general-purpose language models struggle with the precise terminology, unit conventions, and derivations that physics problems demand. We introduce PhysicsMate, a benchmark of 1{,}834 question--answer pairs built from the National Curriculum and Textbook Board (NCTB) Grade~9-10 physics syllabus and grounded in a multi-relational knowledge graph of 1{,}760 nodes and 2{,}600 edges across ten ontological types. We Low-Rank Adapt at 0.6B, 1.7B, and 4B parameters, with a unified recipe and demonstrate a significant increase in closed-book accuracy in all scales (+5.5, +15.0, and +23.3 percentage points).  A node-type analysis shows that the most benefited by adaptation is the structured curricular knowledge, which consists of physical quantities and named laws, while the least benefited is the loosely specified entity-level knowledge. The 4B model has been adapted and quantized to a small offline binary that can be used for local inference in resource constrained environments and offers a viable path to curriculum aligned physics support in environments with limited connectivity and hardware. 
\end{abstract}


\section{Introduction}

Bengali, spoken natively by over 300 million people, remains underrepresented in educational NLP. Existing Bengali benchmarks target sentiment analysis, news classification, and common sense reasoning \cite{bhattacharjee2022banglabert}; and there is a lack of STEM resources in line with the curriculum in secondary education. BEnQA \cite{shafayat2024benqa} and BnMMLU \cite{sarker2025bnmmlu} broaden topical coverage but do not organize questions around a national curriculum, and P6Jiggasha \cite{shahriar2025p6jiggasha} addresses physics one level above the syllabus considered here. This results in a vast gap between the current landscape of language models and classroom needs.

In Bangladesh Secondary physics is based on the NCTB syllabus for grade IX and X which consists of 13 chapters from the topics of mechanics, electromagnetism, optics, and modern physics. Correctness in this domain is not a language- generation problem per se, but involves application of some physical laws, definitions expressed in SI-units, dimensional correctness expressed in the corresponding language, and numerical results expressed in the standard form of the language. General-purpose multilingual models often hallucinate dimensional formulas, generate code-switched texts, and when answering deterministic calculation questions, tend to follow repetitive patterns, a gap that is not just linguistic but ontological – structured internalization of physical relationships is not part of the base models' natural, built-in knowledge. 

There is no existing resource which connects Bengali secondary-physics questions to curriculum structure to facilitate controlled evaluation at the category level.  As illustrated in Fig.~\ref{fig:teaser}, To fill this gap, PhysicsMate provides an end-to-end framework that includes curriculum ingestion, a multi-relational knowledge formalization, a QA benchmark construction, and an adaptation of small models:

\begin{figure*}[t]
\centering
\includegraphics[width=\textwidth]{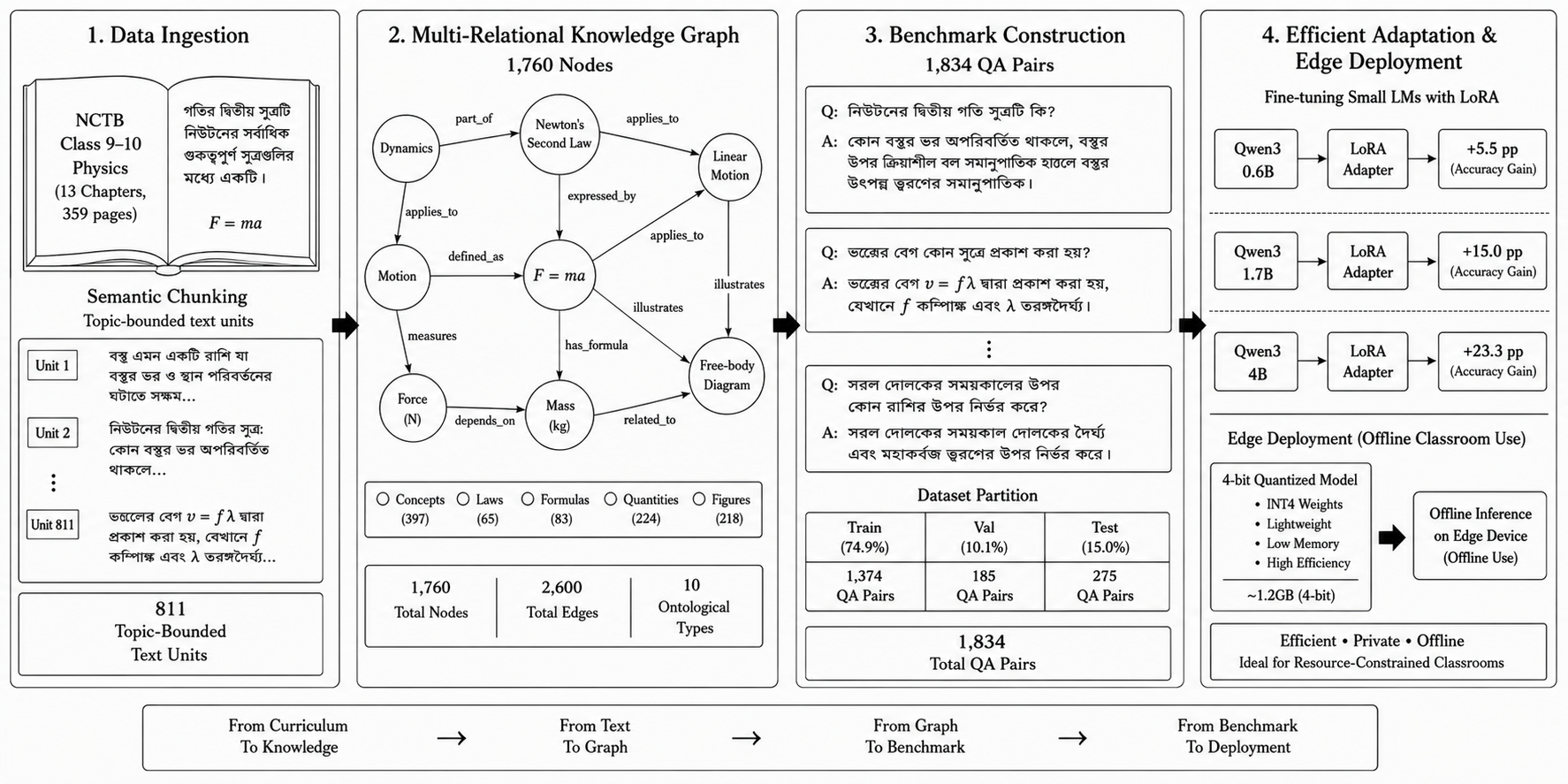}
\caption{The end-to-end PhysicsMate workflow. (1) \textbf{Data Ingestion}: page-level OCR of the NCTB Grade 9-10 physics textbook, Markdown normalization, and three-page contextual knowledge extraction yielding 811 semantic units; (2)\textbf{ Multi-Relational Knowledge Graph}: a 10-type domain ontology with 1,760 nodes and 2,600 directed edges; (3) \textbf{Benchmark Construction}: 1,834 curriculum-grounded Bengali QA pairs stratified into Train/Val/Test; and (4) \textbf{Efficient Adaptation \& Edge Deployment}: LoRA fine-tuning of Qwen3 models with 0.6B, 1.7B, and 4B parameters, along with 4-bit offline quantization. }
\label{fig:teaser}
\end{figure*}

\begin{enumerate}
    \item \textbf{A curriculum-grounded benchmark and knowledge graph.} We formalize the 13-chapter NCTB syllabus as a multi-relational knowledge graph (1{,}760 nodes across 10 ontological types, 2{,}600 directed edges) and use it to construct 1{,}834 QA pairs in formal Bengali, partitioned into train, validation and held-out test sets.
    \item \textbf{A capacity-dependent adaptation study.} Under a single LoRA recipe, we adapt Qwen3 at three scales (0.6B, 1.7B, 4B). As the capacity increases (5.5, 15.0, and 23.3 percentage points accuracy), its gains monotonically increase, but despite the lower gain in accuracy, the adapted model outpaces frontier cloud models on curriculum-aligned lexical and semantic measures. 
    \item \textbf{An ontological diagnostic analysis.} Results broken down by knowledge-node type indicate that structured knowledge like quantities, laws, and definitions adapts most readily, while entity-level knowledge adapts least, providing an empirical basis for targeted improvement in the future.  
\end{enumerate}

\section{Related Work}

\subsection{Bengali Educational and Science QA}

BanglaBERT \cite{bhattacharjee2022banglabert} established a pretrained encoder for Bengali. BEnQA \cite{shafayat2024benqa} introduced parallel Bengali--English science questions and documented a substantial accuracy gap for GPT-4 relative to English. BnMMLU \cite{sarker2025bnmmlu} spans 41 domains with over 134{,}000 multiple-choice questions but provides no ontological structure and does not target secondary physics specifically. P6Jiggasha \cite{shahriar2025p6jiggasha} benchmarks Bangla physics at the Higher Secondary Certificate level using proprietary models, one level above Grade~9-10, without linking questions to curriculum structure. PhysicsMate has been created specially for Grade 9 and 10 Physics and each question is mapped to a curriculum defined ontology.   

\subsection{Efficient Adaptation and Structured QA}

The Qwen3 family \cite{yang2025qwen3} provides dense multilingual models from 0.6B to 32B parameters, and Paramanu \cite{niyogi2026paramanu} shows that compact Bengali-specific models under 400M parameters remain competitive on general tasks. LoRA \cite{hu2021lora} is the standard parameter-efficient adaptation method; Biderman et al.\ \cite{biderman2024lora} report that it generalizes better outside the training distribution than full fine-tuning. On the structured-QA side, KG-for-PhysicsQA \cite{addala2024knowledge} generates knowledge graphs from physics questions for query decomposition, and Christou et al.\ \cite{christou2025ontology} propose an OWL ontology for curriculum materials. Unlike previous systems that construct graphs from text, PhysicsMate utilizes the curriculum-defined ontology to build and partition the benchmark and support the category-level evaluation. 

\section{The PhysicsMate Benchmark}

\subsection{Curriculum Corpus and Semantic Chunking}

PhysicsMate is derived from the official NCTB Grade~9-10 secondary physics textbook, covering 13 chapters and 359 pages (Fig.~\ref{fig:teaser}, Stage~1). We began with page-level optical character recognition (OCR) using Gemini~3.1~Flash~Lite, converting each page into a normalized Markdown representation. In this way formatting noise that has to do with immediate PDF extraction is taken out whilst the design of the text, mathematical expressions, and headings are saved. 

The resulting Markdown pages were then handled using the same model, in three page context windows with 2 page overlaps and a fixed domain-specific persona and structured extraction prompt. The model identified sets of self-contained curriculare knowledge units for each window and associated each with a predefined ontology schema including concepts, physical quantities, figures and examples, entities, definitions, formulas, laws and higher level curricular topics. Offering three consecutive pages at once as local context enabled knowledge units to be captured that spanned across pages. We ensured that each unit kept its topic association and that each one kept its source of origin (in the textbooks) per knowledge record so every one is traceable back to the originating text. 

The records were subsequently normalized and duplicated and arranged according to the boundaries of topics into the corpus of 811 semantic units. In contrast to fixed-length segmentation, each is formulated so that a full curricular concept, possible mathematical notation, the conventions for SI- units, full worked examples, and figure references are preserved.  These structured records were used directly as input for the knowledge graph creation and benchmark generation. 

\subsection{Knowledge Graph Architecture}

We used a pre-defined ontology for the secondary physics domain to represent the extracted curricular units as a multi-relational knowledge graph  (Fig.~\ref{fig:teaser}, Stage~2). The ontology represents complementary types of curricular knowledge, such as concepts, questions, physical quantities, figures, examples, entities, definitions, formulas, laws and higher level curricular topics. Each knowledge record was associated with the source of where it came from, thus providing source traceability from graph nodes back to the textbook. 

Based on the semantic role of each record extracted, relations were assigned. The principal relations used are those of curricular hierarchy  (\texttt{part\_of}), applicability of laws and formulas to physical situations (\texttt{applies\_to}), visual grounding of figures and concepts  (\texttt{illustrates}), example relations (\texttt{example\_of}), dependency relations (\texttt{depends\_on}), and provenance relations such as \texttt{derived\_from}. This representation is a structured curricular index of the source material and is utilized for the category level analysis of the benchmark. 

The resulting graph contains 1{,}760 nodes across 10 ontological types and 2{,}600 directed edges spanning 30 relation types. Table~\ref{tab:kg_stats} summarizes the node and relation statistics.

\begin{table}[t]
\caption{Knowledge graph statistics. Node types correspond to four cognitive levels: declarative recall (definitions, laws), procedural application (formulas, examples), visual interpretation (figures), and curricular scaffolding (HSC topics).}
\label{tab:kg_stats}
\centering
\footnotesize
\setlength{\tabcolsep}{3pt}
\begin{tabular}{lrlr}
\toprule
\textbf{Node Type} & \textbf{Count} & \textbf{Top Edge Type} & \textbf{Count} \\
\midrule
Concept    & 397 & \texttt{part\_of}      & 585 \\
Question   & 252 & \texttt{applies\_to}   & 531 \\
Quantity   & 224 & \texttt{illustrates}   & 361 \\
Figure     & 218 & \texttt{example\_of}   & 268 \\
Example    & 188 & \texttt{depends\_on}   & 215 \\
Entity     & 176 & \texttt{derived\_from} & 118 \\
Definition & 136 & \multicolumn{2}{l}{\textit{+ 24 further relation types}} \\
Formula    &  83 & & \\
Law        &  65 & & \\
HSC Topic  &  21 & & \\
\midrule
\textbf{Total nodes} & \textbf{1{,}760} & \textbf{Total edges} & \textbf{2{,}600} \\
\bottomrule
\end{tabular}
\end{table}

To make the extraction procedure reproducible, Table~\ref{tab:node_schema} lists the fields that each knowledge record carries. Each record has been typed following the ontology, and has a self-contained Bengali description of the curricular item, with links to the source pages from which it was extracted.   Mathematical expressions, physical units, and typed relations to other nodes are stored alongside each record, giving the graph both factual content and explicit relational structure.

\begin{table}[t]
\caption{Fields stored for each extracted knowledge record. Every record is assigned one ontology type and retains full provenance back to the source textbook pages.}
\label{tab:node_schema}
\centering
\footnotesize
\setlength{\tabcolsep}{3pt}
\begin{tabularx}{\columnwidth}{lX}
\toprule
\textbf{Field} & \textbf{Description} \\
\midrule
\texttt{node\_type} & Ontology category (concept, law, formula, etc.). \\
\texttt{title} & Canonical name of the curricular item. \\
\texttt{content} & Self-contained Bengali description of the knowledge unit. \\
\texttt{formula} & Mathematical expression, when the item includes one. \\
\texttt{unit} & SI or textbook-specified physical unit, where applicable. \\
\texttt{source\_pages} & Textbook page identifiers used as evidence. \\
\texttt{topic} & Chapter and topic within the NCTB curriculum. \\
\texttt{relations} & Typed links to related nodes (concepts, laws, figures, etc.). \\
\texttt{provenance} & Reference to the semantic unit of origin. \\
\bottomrule
\end{tabularx}
\end{table}

\subsection{QA Construction and Partitioning}

The knowledge items and the source contexts used for benchmark items were structured.  (Fig.~\ref{fig:teaser}, Stage~3). Each question-answer pair is linked to a specific knowledge node and originating semantic unit (curricular provenance): each question-answer pair can be traced back to a particular passage of the textbook, its representation as an ontology-typed semantic unit. The questions cover the entire physics syllabus for class IX-X of the National Council for Education Research Training (NCTB) and are written in the standard Bengali language. 

The final benchmark contains 1{,}834 question--answer pairs. Chapter~1 (Physical Quantities and Measurement) contributes the most items (302), while Chapter~12 (Magnetic Effects of Current) contributes the fewest (82). The corpus is partitioned into training ($n=1{,}374$; 74.9\%), validation ($n=185$; 10.1\%), and held-out test ($n=275$; 15.0\%) sets, stratified by chapter. Partition membership was checked programatically to ensure that no question ID is listed in more than one partition. The average question length is 50.3 characters and the average answer length is 117.2 characters.

\subsection{Data Provenance and Quality Control}

The key design decision of PhysicsMate is that the NCTB textbook is the source of truth for all content. The extraction pipeline does not produce questions based on general knowledge, but rather, each benchmark item is based on a specific passage in a text book, which is kept as metadata throughout the pipeline. This means that each answer should be presented in the terminology, formulas and conventions of the units of the source curriculum, not in paraphrases taken from other curricula.  

The fixed schema and structured prompt in the knowledge extraction process limit the variability of the representation of equivalent curricular facts: the same physical law will always be represented as its name designated by the NCTB and its formula, not as a variant of the formula. The partition assignment was done after the entire benchmark was developed and each question identifier was compared across all three partitions to ensure there was no overlap. These properties enable fine tuning and evaluation to stay clearly separated and reproducible. 

\subsection{Dataset Release and Documentation}

The multi-relational knowledge graph and all 1,834 QA pairs are released upon publication at \url{https://github.com/jahin-7/physicsmate}
, accompanied by a Dataset Card documenting the extraction pipeline, ontology schema, data provenance, and license. The released splits are identical to those used for evaluation here, so no additional partitioning is needed to reproduce our results; each record's metadata links it back to its textbook chapter, node type, and any mathematical expressions it introduces.

\section{Experimental Setup}

\subsection{Models and Adaptation}

We evaluate three Qwen3 scales: 0.6B, 1.7B, and 4B parameters. Each model is adapted using LoRA \cite{hu2021lora}, applied to all seven linear projection layers per transformer block (query, key, value, output, gate, up, down), with rank $r=16$, scaling factor $\alpha=32$, and dropout $p=0.05$, held fixed across all three scales. Models are trained for 3 epochs with 8-bit AdamW at a learning rate of $2\times10^{-4}$ under cosine decay, with base weights loaded in 4-bit NF4 quantization; only the LoRA adapters are updated, approximately 0.8--1.7\% of total parameters depending on scale. Adaptation recipe is the same for every scale so the key variable is the capacity of the models in the three situations. 

\subsection{Evaluation}

All models are evaluated in closed-book mode on the 275 held-out test questions with greedy decoding, capped at 160 new tokens. We report three metrics: Token F1, a lexical overlap measure using a Bengali Unicode regex tokenizer; multilingual BERTScore \cite{zhang2019bertscore}; and LLM-as-a-Judge accuracy \cite{zheng2023judging}. The judge, Gemini-2.5-Flash-Lite at $T=0$, is given the question, gold answer, and prediction (all in Bengali) and returns a single token, \textsc{correct} or \textsc{incorrect}, marking a prediction correct only if it conveys the gold reference's key fact; the identical prompt and protocol are applied to the edge models and the zero-shot cloud baselines alike (GPT-4o-mini, Gemini-2.5-Flash-Lite). The full prompt is released with the code. We did not cross-validate this judge against an independent model or a human-annotated subset; we flag this explicitly as an open limitation (Section~\ref{sec:discussion}). 

\section{Results}
\label{sec:results}

\subsection{Main Benchmark}

Table~\ref{tab:main_results} reports results across all conditions. Fine-tuning improves every metric at every scale.

\begin{table}[t]
\caption{Results on the held-out test set ($n=275$). Accuracy is LLM-judge (Gemini-2.5-Flash-Lite). Cloud baselines are zero-shot; edge models are evaluated with and without LoRA adaptation.}
\label{tab:main_results}
\centering
\footnotesize
\setlength{\tabcolsep}{1.8pt}
\begin{tabular}{llcccccc}
\toprule
\textbf{Model} & \textbf{Params} & \multicolumn{2}{c}{\textbf{Acc.\ (\%)}} & \multicolumn{2}{c}{\textbf{Token F1}} & \multicolumn{2}{c}{\textbf{BERTScore}} \\
\cmidrule(lr){3-4} \cmidrule(lr){5-6} \cmidrule(lr){7-8}
& & Raw & FT & Raw & FT & Raw & FT \\
\midrule
\multicolumn{8}{l}{\textit{Cloud baselines (zero-shot)}} \\
GPT-4o-mini & $>$8B & \multicolumn{2}{c}{69.5} & \multicolumn{2}{c}{0.239} & \multicolumn{2}{c}{0.728} \\
Gemini-2.5-FL & $>$10B & \multicolumn{2}{c}{82.9} & \multicolumn{2}{c}{0.267} & \multicolumn{2}{c}{0.733} \\
\midrule
\multicolumn{8}{l}{\textit{Edge models (local, offline)}} \\
Qwen3-0.6B & 0.6B & 0.0 & \textbf{5.5} & 0.216 & \textbf{0.278} & 0.696 & \textbf{0.750} \\
Qwen3-1.7B & 1.7B & 10.5 & \textbf{25.5} & 0.227 & \textbf{0.299} & 0.706 & \textbf{0.765} \\
Qwen3-4B & 4.0B & 27.6 & \textbf{50.9} & 0.252 & \textbf{0.350} & 0.722 & \textbf{0.782} \\
\bottomrule
\end{tabular}
\end{table}

The base model, Qwen3-0.6B, is not able to get the correct answer rate on the test set, but following adaptation, the rate becomes 5.5\%, which suggests that even a sub-billion-parameter model gets some domain grounding. The accuracy of the 1.7B model is doubled, from 10.5\% to 25.5\%, and the edge performance of the 4B model is 23.3 percentage points higher than their respective baseline model. Similarly, the relative improvements in Token F1 range from 28.3\% at 0.6B to 31.5\% at 1.7B and 38.9\% at 4B. 

Cloud baselines have greater raw accuracy, 69.5 and 82.9 percent respectively but lower Token F1 and BERTScore than the adapted 4B model. The nature of the gap is as follows: The cloud models are more likely to generate longer and more explanatory responses compared to the shorter NCTB-standard phrases that the lexical and semantic metrics judge even in cases when the reasoning underpinning the response is well justified. 

\subsection{Capacity-Dependent Adaptation}
\label{sec:capacity}

\begin{figure}[t]
\centering
\includegraphics[width=\columnwidth]{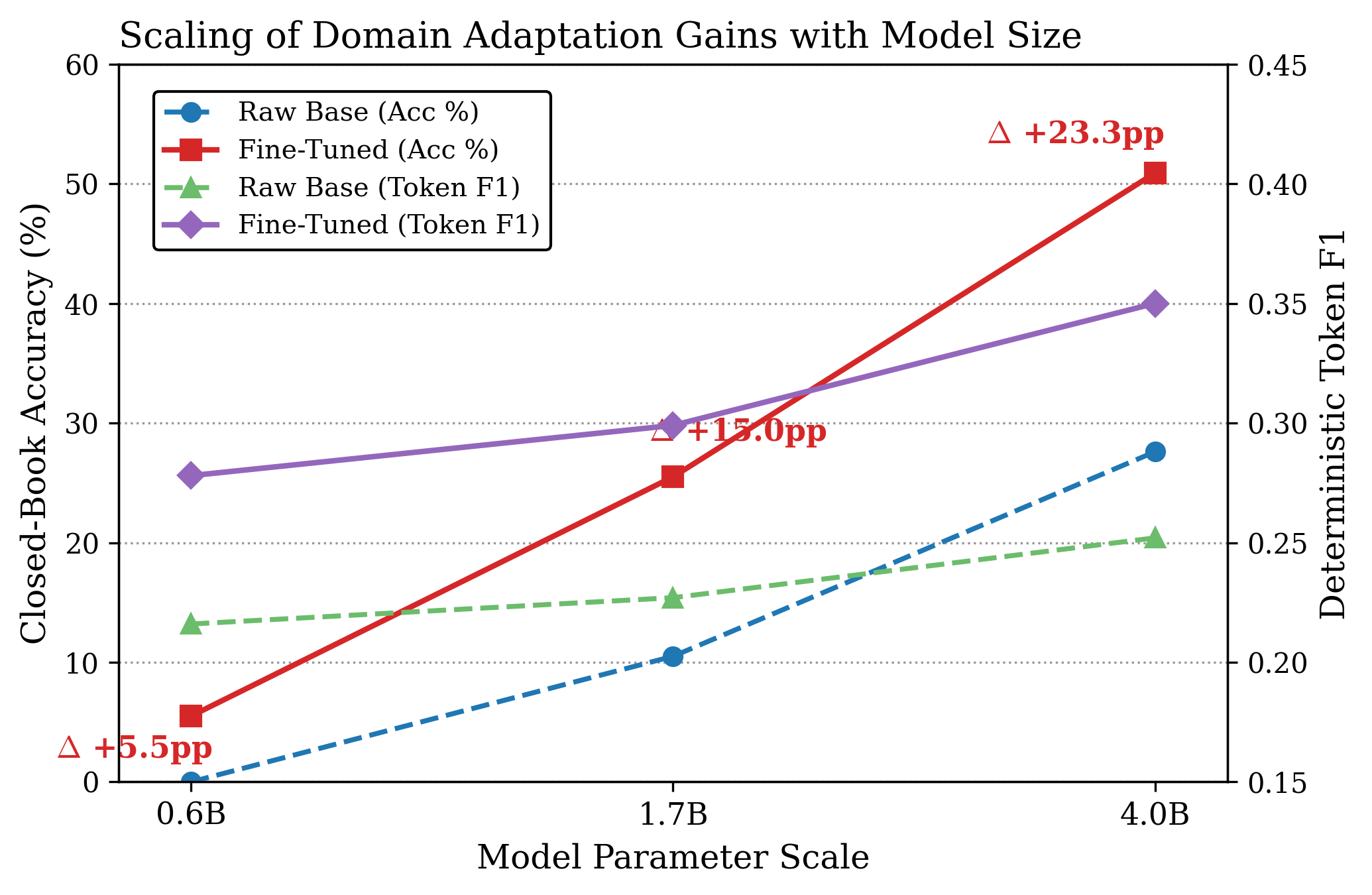}
\caption{Capacity-dependent adaptation gains. Fine-tuning produces monotonically increasing accuracy improvements (+5.5, +15.0, and +23.3 percentage points) and consistent gains in Token F1 and BERTScore as model size grows from 0.6B to 4B parameters.}
\label{fig:scaling}
\end{figure}

Figure~\ref{fig:scaling} shows that adaptation gains grow with model capacity. The accuracy improvement trajectory is:
\begin{equation}
\Delta\text{Acc}: \; +5.5\text{\,pp} \;\to\; +15.0\text{\,pp} \;\to\; +23.3\text{\,pp}
\end{equation}
The results suggest that bigger base models leverage the same LoRA capacity more effectively to encode domain-specific structure, with an improvement of 50.9\% (from 25.5\% to 50.4\%) in fine-tuned accuracy from 1.7B to 4B. Three data points is not enough to prove a scaling law, but this trend is observed across each of the three metrics—accuracy, Token F1, and BERTScore—takes a single glance at which you can see it is monotonic. 

\subsection{Node-Type Diagnostics}

Table~\ref{tab:node_f1} breaks down Qwen3-4B Token F1 by knowledge-node type on the test set. Quantity nodes show the largest relative gain (93.9\%), followed by scientific laws (56.6\%) and definitions (56.1\%). Entity nodes show the smallest gain (8.7\%).

\begin{table}[t]
\caption{Qwen3-4B Token F1 by knowledge-node type ($n=275$ test questions). Structured knowledge types benefit most from adaptation.}
\label{tab:node_f1}
\centering
\footnotesize
\setlength{\tabcolsep}{3pt}
\begin{tabular}{lcccc}
\toprule
\textbf{Node Type} & $n$ & \textbf{Raw F1} & \textbf{FT F1} & \textbf{Rel.\ Gain} \\
\midrule
Quantity   & 29 & 0.262 & \textbf{0.508} & 93.9\% \\
Law        &  8 & 0.286 & \textbf{0.448} & 56.6\% \\
Definition & 28 & 0.228 & \textbf{0.356} & 56.1\% \\
Example    & 35 & 0.244 & \textbf{0.358} & 46.7\% \\
Concept    & 65 & 0.250 & \textbf{0.342} & 36.8\% \\
Figure     & 67 & 0.263 & \textbf{0.320} & 21.7\% \\
Formula    &  8 & 0.280 & \textbf{0.332} & 18.6\% \\
Entity     & 35 & 0.241 & \textbf{0.262} & 8.7\% \\
\bottomrule
\end{tabular}
\end{table}

Ordering is interpretable: quantities, laws, and defini- tions are concrete, well-structured facts with canonical Bengali phrasings in the NCTB textbook, and entity mentions are loosely specified, and often admit multiple valid surface forms. The former is captured better by low-rank adaptation than the latter, which follows the pattern seen in Section ~\ref{sec:capacity}  as a function of capacity. 

\begin{figure}[t]
\centering
\includegraphics[width=\columnwidth]{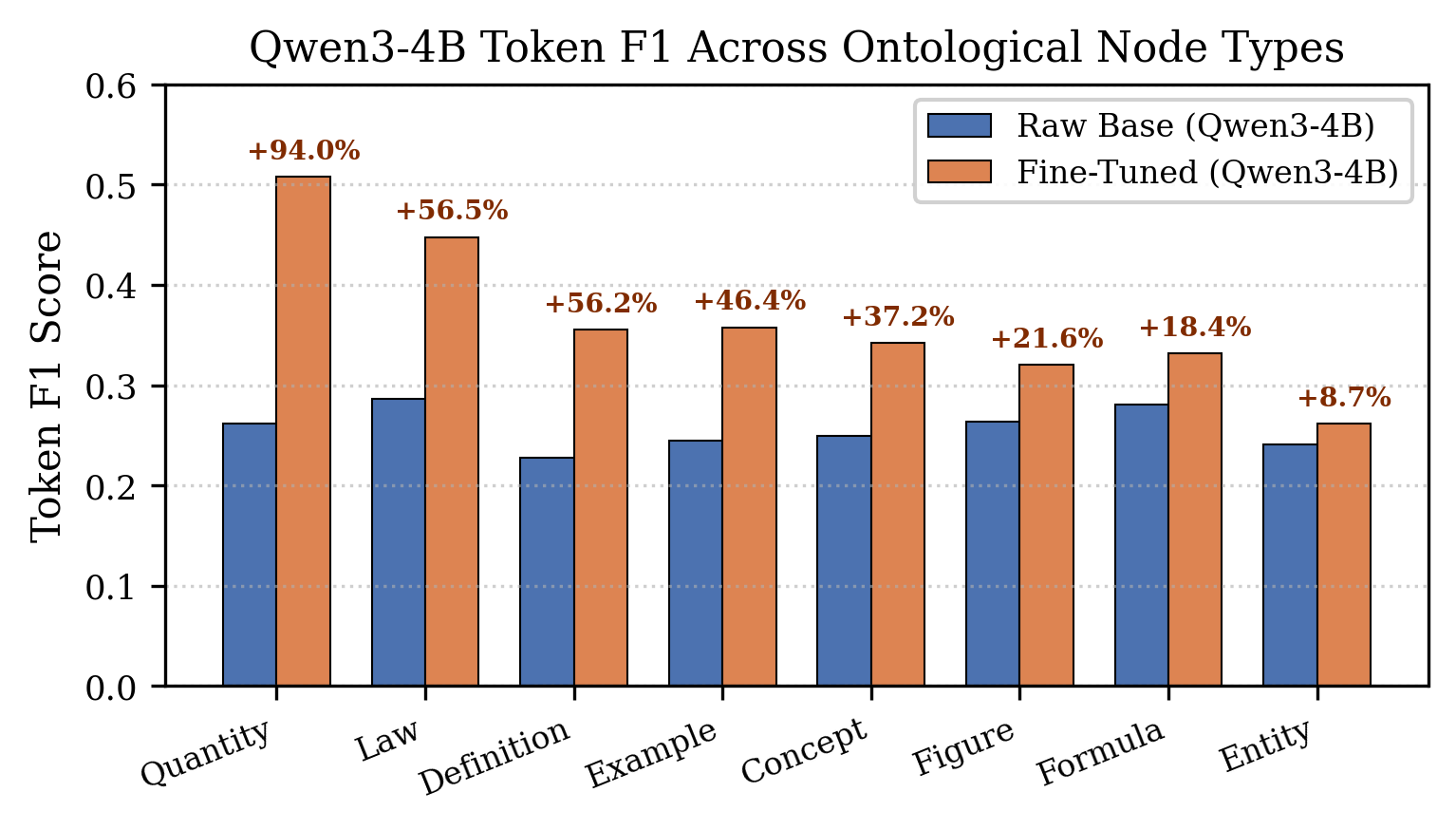}
\caption{Token F1 by knowledge-node type for base vs.\ fine-tuned Qwen3-4B. The largest gains appear on structured physical quantities (93.9\%) and scientific laws (56.6\%); entity-level knowledge shows the smallest improvement (8.7\%).}
\label{fig:node_f1}
\end{figure}

\subsection{Qualitative Analysis}

Table~\ref{tab:qual} presents two cases illustrating failure modes that adaptation corrects.

\begin{table*}[t]
\caption{Representative base vs.\ fine-tuned Qwen3-4B outputs. Fine-tuning corrects factual hallucination and incomplete derivation.}
\label{tab:qual}
\centering
\scriptsize
\setlength{\tabcolsep}{3pt}
\renewcommand{\arraystretch}{1.15}
\begin{tabularx}{\textwidth}{p{2.8cm} p{3.4cm} p{4.2cm} X}
\toprule
\textbf{Category} & \textbf{Gold Reference} & \textbf{Base Qwen3-4B} & \textbf{Fine-Tuned Qwen3-4B} \\
\midrule
Optics (ch09\_q084)
\newline \textit{Image properties for an object between a convex lens and its focus?}
& \bn{প্রতিবিম্ব অবাস্তব, সোজা ও বস্তুর তুলনায় বড়ো হয়।}
\newline \textit{[Virtual, erect, magnified.]}
& \textit{Fabricated rule:} \bn{প্রতিবিম্ব বামে দাঁড়ায় (বস্তুর বাম দিকে)...}
\newline \textit{[Invents a spatial rule absent from the curriculum.]}
& \bn{প্রতিবিম্ব অবাস্তব, সোজা ও বস্তুর তুলনায় বড় হয়।}
\newline \textit{[Matches gold reference.]}
\\ \addlinespace
Circuits (ch11\_q031)
\newline \textit{Express power in terms of current and resistance.}
& $P = I^2 R$
\newline \textit{[Power dissipated as heat in the resistor.]}
& \textit{Incomplete:} \bn{$V = I \times R$, যেখানে $V=$ ভোল্ট, $I=$ প্রবাহ...}
\newline \textit{[Stops at Ohm's law; never derives the power expression.]}
& \bn{$P = I^2 R$ বা $P = V^2/R$ দ্বারা ক্ষমতাকে প্রকাশ করা যায়।}
\\
\bottomrule
\end{tabularx}
\end{table*}

The base model of the optics case generates a spatial rule that is not included in the curriculum and the adapted model generates the correct answer of three adjectives. In the circuits case, the base model reads out Ohm's Law but never actually does the algebraic step to the power formula; the adapted model simply reads out $P=I^2R$. Both patterns agree with the node- type results above: adaptation is best in places where there is one canonical answer. 

\section{Discussion, Limitations, and Conclusion}
\label{sec:discussion}

These experiments have led to 3 observations. Second, parameter-efficient adaptation is materially beneficial at all scales of models tested, including one that generates no correct answers without adaptation, for Bengali secondary-physics QA. Second, the extent of improvement is heavily dependent on capacity: With the same LoRA configuration, we achieve an improvement of 23.3 percentage points at 4B and 5.5 percentage points at 0.6B, indicating that larger models offer more useful representational space for a given domain structure.  Third, the improvements are not distributed uniformly over the categories of curricular knowledge, but are focused on well-structured categories like quantities, laws, and definitions in which the knowledge can be expressed in a canonical way, and are less for loosely specified entity knowledge. 

The adapted 4B model is not anywhere close to frontier-model accuracy (50.9\% vs. 82.9\%) and we don't say that PhysicsMate solves Bengali physics QA. What the results show is that curriculum-based adaptation works, and that the benchmark offers the diagnostic detail necessary to determine where more work needs to be done. The 4B model is quantized to an offline binary version that can be deployed for local inference, which is important for applications where connectivity is limited. 

Although PhysicsMate is limited to the specific context of the NCTB Grade 9-10 Physics textbook, the multirelational extraction framework is designed to be generalizable in two ways: both within adjacent disciplines and in neighboring educational systems. The 10-node ontology is directly isomorphic to a domain in the NCTB secondary framework: secondary Chemistry corresponds directly to the domain of “physical quantities”, “named laws” (e.g., gas laws), “formulae”, and “definitions”, though a few changes have to be made to incorporate chemical reactions and compound entities; secondary Biology corresponds to “hierarchical concepts”, “anatomical figures”, “taxonomic entities”, and “descriptive definitions”. The declarative and procedural primitives would have to be restructured into causal, temporal and narrative relation types for non-STEM or humanities curricula (e.g., Bangladesh and Global Studies). PhysicsMate is an educational framework across national boundaries which lays the groundwork for secondary education in South Asia. Secondary science curricula throughout South Asia, such as NCERT/CBSE in India, West Bengal Board of Secondary Education (WBBSE) and Federal Board in Pakistan are aligned to classical physics at Grade 9-10 with a common system of units (SI units). This approach is based on structural relations, not language-specific idioms, so our ingestion and ontology schemas can be directly applied to bilingual or vernacular contexts throughout South Asia (e.g., West Bengal Bengali, Hindi, Urdu) and can serve as an empirical blueprint for cross-curriculum benchmarking in under-resourced regional education ecosystems. 

Several limitations apply. The evaluation is text-only; the 218 figures linked in the knowledge graph are not used as visual inputs. Accuracy is based on LLM judge, not human annotation. One of the two zero-shot cloud baselines reported in Table~\ref{tab:main_results} is Gemini-2.5-Flash-Lite, which is self-assessed. The accuracy of this judge was not cross-checked against an independent judge model or a human-annotated subset. The held-out split is disjoint by question ID, though we do not claim a broader contamination analysis beyond that check. The benchmark covers a single national curriculum, NCTB Grade~9-10 physics, and generalization to other syllabi or subjects has not been tested.

Curriculum extraction, benchmark synthesis and edge deployment relying on LLM bring key pedagogical and ethical issues. Automated knowledge extraction with proprietary LLMs may lead to "silent hallucinations," "ontological omissions," or nuanced scientific inaccuracies in the benchmark that are not thoroughly human-checked. In secondary education learners are at the stage of having no domain discernment and if they are exposed to realistic but factually incorrect concepts like spatial optic hallucinations or incomplete circuit derivations in unadapted models, misconceptions may be reinforced. Secondly, parameter-efficient adaptation leads to higher curricular alignment, but there is still a lot of room to error in our best offline model with 50.9\% closed book accuracy. In this regard, it may be dangerous to deploy such models directly to students in unmoderated resource-constrained classroom environments without any teacher in the loop framework. Thirdly, the construction of synthetic benchmarks should be done in such a way to avoid stylistic and linguistic homogenisation: The automated QA pairs may be the idiosyncratic lexical distributions of the extractor model and may not necessarily represent the authentic dialectal, pedagogical and enquiry patterns of Bengali classroom discourse. Two additional notes about the pipeline: First, there is no risk of misconception being reinforced because no student information is sent to the pipeline; second, although the deployment is offline, this does not eliminate the resource gap, since the pipeline requires the presence of at least one device capable of handling the content, regardless, a classroom must have some devices with content capabilities. The source content is a nationally mandated, copyrighted textbook, and any public release of the derived knowledge graph and QA pairs should include a clear provenance statement, and adhere to the licensing terms of NCTB.  Thus, PhysicsMate and models derived from it should be placed as a tool to assist teaching, reviewed by a teacher rather than deployed as an autonomous grader, and a yardstick for evaluation of learning, more than a pedagogically independent entity. 

PhysicsMate offers a curriculum-aligned Bengali secondary physics benchmark, a controlled adaptation study spanning three model scales, and an ontological diagnostic to pinpoint where parameter-efficient adaptation can be most beneficial. These components provide an empirical basis for the design and assessment of efficient models of educational language for Bengali. 

\bibliographystyle{IEEEtran}
\bibliography{references}

\end{document}